\documentclass[11pt,a4paper]{article}
\usepackage[hyperref]{acl2018}
\usepackage{times}
\usepackage{latexsym}
\usepackage{amsmath}   
\usepackage{graphicx}
\usepackage{makecell}
\usepackage{microtype}

\usepackage{url}

\aclfinalcopy 
\def\aclpaperid{***} 

\title{Spelling Error Correction Using a Nested RNN Model and Pseudo Training Data}

 \author{Hao Li$^1$ \and Yang Wang$^1$ \and Xinyu Liu$^2$ \and Zhichao Sheng$^1$ \and Si Wei$^1$ \\
      \{haoli5,yangwang2,zcsheng,siwei\}@iflytek.com\\
      liuxy.zoe2016@outlook.com\\\\
             $^1$iFLYTEK Research, Hefei, China \\$^2$International Department, The Affliated School of SCNU, Guangzhou, China}
\date{}

\begin{document}
\maketitle
\begin{abstract}
    We propose a nested recurrent neural network (nested RNN) model for English spelling error correction and generate pseudo data based on phonetic similarity to train it.
    The model fuses orthographic information and context as a whole and is trained in an end-to-end fashion.
    This avoids feature engineering and does not rely on a noisy channel model as in traditional methods.
    Experiments show that the proposed method is superior to existing systems in correcting spelling errors.
\end{abstract}

\section{Introduction}
Spelling error correction ~\cite{Kukich:1992, jurafsky:2000} is a basic issue in grammatical error correction (GEC) and is widely used in practical scenarios ~\cite{PubMed:2006}.
Most traditional systems in this field are based on Levenshtein Distance and statistical methods ~\cite{Damerau:1964, Kashyap:1983, MDM:1990, Kemighan:1990, Toutanova, MDM:2008}.
Despite being successful in GEC, neural network models have not drawn much attention in correcting spelling errors.

Spelling errors are often divided into two categories: non-word errors and real-word errors ~\cite{jurafsky:2000}.
Non-word spelling errors can be detected via a dictionary.
Any word that is not in a dictionary is an error.
Real word spelling errors are much more difficult to identify since the misspelled words are in the vocabulary ~\cite{MDM:1990, MDM:2008}.

Most traditional systems for spelling error correction are based on the noisy channel model, where a true word is distorted as if passed through a noisy communication channel ~\cite{Kemighan:1990, Brill:2001}.


Recently, GEC becomes an active field ~\cite{conll:2014} and data-driven methods are proposed against spelling errors.
~\citet{chollampatt-ng:2017:BEA} proposed character-level statistical machine translation to ``translate'' unknown words into correct words.
~\citet{scRNN:2017:AAAI} proposed scRNN based on psycholinguistics experiments. 
In many neural network based GEC systems, words are split into character level or BPE level to avoid OOV words~\cite{schmaltz:2016:BEA11, ji-EtAl:2017:Long, cnngec}.
These models consider spelling errors and grammatical errors as a whole.
But their performance on spelling error correction is limited due to a lack of training data in GEC ~\cite{dahlmeier-ng-wu:2013:BEA8}.

In this paper, we propose a nested RNN model for spelling error correction and train it with pseudo data.
The contributions are listed as follows.

\begin{itemize}
  \item We propose a stand-alone model for spelling error correction, where both orthographic information and information are considered.
  \item We generate pseudo data based on phonetic similarity to train the model.
\end{itemize}

\section{Related work}

Traditional spelling correction methods mostly depends on the noisy channel model ~\cite{Kemighan:1990, Church:1991}.
~\citet{Brill:2001} improves it by splitting a word into partitions.
~\citet{Toutanova} incorporates pronunciation with a letter-to-sound model.

Language models are also used to detect semantic anomaly based on the context, especially for dealing with real-word errors ~\cite{MDM:1990, MDM:2008}.
But the probabilities from noisy channel and language model are often not commensurate.
Besides, Word-Net are also used to detect potential corrections~\cite{Hirst:2005}.

Neural networks are also adapted.
~\citet{scRNN:2017:AAAI} proposed the scRNN model by ignoring the orders of internal characters of a word.
But the model actually lost much orthographic information.
On the other hand, in many GEC systems, OOV words are split in character-based or BPE-based units and fed into sequence-to-sequence models ~\cite{xieziang, schmaltz:2016:BEA11, ji-EtAl:2017:Long, cnngec}.
By doing this, the task of spelling error correction is fused into GEC and the two tasks are trained end-to-end together.
However, in the field of GEC, these systems suffer from a lack of labeled data ~\cite{dahlmeier-ng-wu:2013:BEA8} and are not trained enough against spelling errors.
Hence, other GEC systems would rather use traditional stand-alone methods, such as enchant\footnote{https://github.com/AbiWord/enchant}, to correct the spelling errors beforehand \cite{RL:2017, napoles-callisonburch:2017:BEA}.

Compared to the methods above, the nested RNN model is novel.
First, it combines orthographic information and context and is trained in an end-to-end fashion, avoiding parameter tuning.
Second, it can be trained with large-scale pseudo data, without using human-generated resources.

\section{the Nested RNN model}

\begin{figure}[ht]
    \centering
    \includegraphics[scale=0.32]{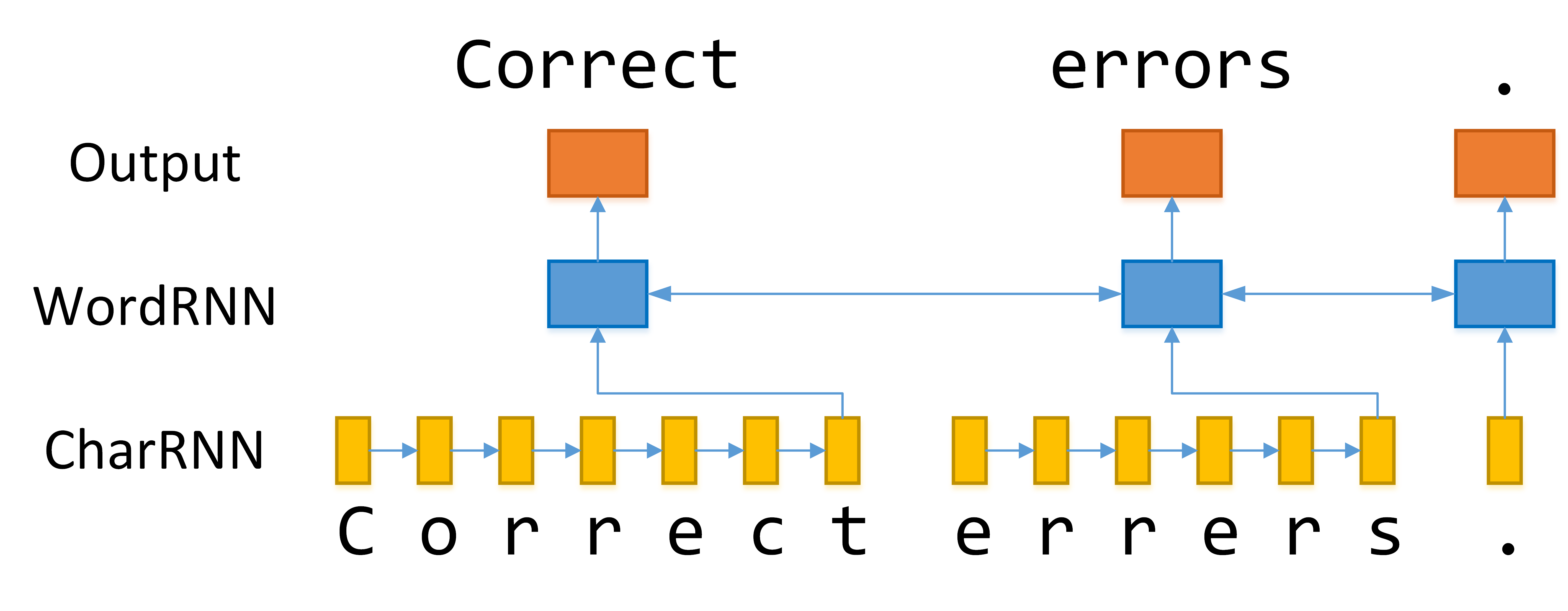}
    \caption{Architecture of the Nested RNN Model}
    \label{architecture}
\end{figure}

We convert spelling error correction into a sequence labeling problem, where a correct word is labeled as itself and a misspelled word is labeled as its corrected version.
The model is shown in Fig~\ref{architecture}.
It consists of a char-level RNN (CharRNN) and a word-level one (WordRNN).
The CharRNN collects orthographic information by reading each word as a sequence of letters.
The WordRNN predicts the correct words by combining the orthographic information with the context.

\subsection{CharRNN}
Given a word $w$, we map its character sequence $(c_{1}, c_{2}, \dots, c_{L})$ into a series of character embeddings $\mathbf{e}$:
\begin{equation}
    \mathbf{e} = (e_{1}, e_{2}, \dots, e_{L})
\end{equation}
The CharRNN then encodes $\mathbf{e}$ into a sequence of hidden vectors $(s_1, s_2, \dots, s_L)$. where the $l$th hidden state $s_l$ is computed as:
\begin{equation}
    s_l = \textrm{GRU}_{\textrm{c}}(s_{l-1}, e_l)
\end{equation}
and GRU is the gated recurrent unit ~\cite{gru}, a modified version of the vanilla RNN.

We take the last hidden vector $s_L$, denoted as $\bar{s}$, as a representation of $w$.
Thus, the CharRNN encodes a sentence with words $(w_1, w_2, \dots, w_n)$ into a sequence of vectors $\bar{\mathbf{s}}$:
\begin{equation}
    \bar{\mathbf{s}} = (\bar{s}_1, \bar{s}_2, \dots, \bar{s}_N)
\end{equation}

\subsection{WordRNN}
For the sentence above, the WordRNN takes $\bar{\mathbf{s}}$ as input and creates a corresponding sequence of vectors $\mathbf{h}$:
\begin{equation}
    \mathbf{h} = (h_{1}, h_{2}, \dots, h_{N})
\end{equation}
The $n$th hidden state $h_n$ is computed as
\begin{eqnarray}
    h_n &=& [h^{\textrm{f}}_n; h^{\textrm{b}}_n]  \\
    h^{\textrm{f}}_n &=& \textrm{GRU}^{\textrm{f}}_{\textrm{w}}(h^{\textrm{f}}_{n-1}, \bar{s}_n) \\
    h^{\textrm{b}}_n &=& \textrm{GRU}^{\textrm{b}}_{\textrm{w}}(h^{\textrm{b}}_{n+1}, \bar{s}_n)
\end{eqnarray}
where $\textrm{GRU}^{\textrm{f}}_{\textrm{w}}$ and $\textrm{GRU}^{\textrm{b}}_{\textrm{w}}$ denote the forward and backward word-level GRU cell, respectively.

Then the probability of each target word $y_n$ is computed as:
\begin{equation}
    p(y_n|\mathbf{e}) = \mathrm{softmax}(g(h_n))
\end{equation}
where $g$ is a linear function that maps the hidden state into a vector of target vocabulary size.

The model is trained by minimizing the cross-entropy loss, which for a given $(\mathbf{x},\mathbf{y})$ pair is
\begin{equation}
    loss(\mathbf{x},\mathbf{y}) = -\sum_{n=1}^{N}{\log p(y_n|\mathbf{x})}
\end{equation}

\subsection{Processing unknown words}
The nested RNN model has a limited output vocabulary size.
The out-of-vocabulary tokens in the output are represented by a special UNK symbol.
Thus post-processing on the UNKs is needed for inference.
Noticing that for our spelling correction model, the input and output words are matched one-by-one, we simply replace all the UNKs in the output with the corresponding source word.

\section{Training with pseudo data}
To train the nested RNN model, large quantity of labeled data is needed, which is not available to our knowledge.
So we propose to train it with pseudo generated data.

Words are often spelled in the way they sound like, which means, a word is more easily misspelled to those with similar pronunciation, e.g. $\textit{understand} \to \textit{understend}$, $\textit{decision} \to \textit{dicision}$.
This property has been used for improving the noisy channel model ~\cite{Toutanova}.

Inspired by this, we propose a simple method to generate pseudo training data for spelling error correction based on phonetic similarity.

\subsection{Character-level confusion matrix}
First, we collect words from a dictionary $\mathcal{D}$ and train an attention-based NMT ~\cite{Bahdanau:2015} model from word to pronunciation, which, for a single word, takes its sequence of characters as input and the corresponding sequence of phonetic symbols as output.
Then for each word $w$ in $\mathcal{D}$ , we calculate the normalized attention weights between phonetic symbols $p$ and characters $c$:
\begin{equation}
    {Att}(w, c_l, p_k), w \in \mathcal{D}, 1 \leq l \leq L , 1 \leq k \leq K
\end{equation}
where $L$ and $K$ are the number of characters and phonetic symbols in word $w$.

We consider $c$ and $p$ to be ``aligned'' in $w$ if $Att(w,c,p)$ exceeds some threshold $\theta_{att}$. If two or more adjacent characters are aligned by a same phonetic symbol, they will be merged to form a new ``character''. So in this chapter, the word ``character'' may refer to a real character or a sequence of characters. For example, from the word \textit{beam} (/bi:m/), we may get the alignment pairs (\textit{b}, b), (\textit{ea}, i:), (\textit{m}, m).

Then for each character $c$ and phonetic symbol $p$ in the alphabet, we count the number of times they are aligned in the dictionary:
\begin{eqnarray}
    & N(c,p) = \sum_{w \in \mathcal{D}}{align(w, c, p)}
\end{eqnarray}
where
\begin{equation}
    align(w, c, p) =
    \left\{
        \begin{array}{ll}
            1, & {Att}(w,c,p) > \theta_{att}   \\
            0, & \textrm{otherwise}
        \end{array}
    \right.
\end{equation}
For ordered character pair $(c_1, c_2)$, define the phonetic confusion coefficient as:
\begin{equation}
    M(c_1, c_2) = \frac{\sum_{p}{N(c_1,p) \cdot N(c_2,p)}}{\sum_{\hat{c} \in \mathcal{D}_p}{\sum_{p}{N(c_1,p) \cdot N(\hat{c},p)}}}
\end{equation}
where $\mathcal{D}_p$ is the phonetic alphabet.

Character pairs with large confusion coefficients tend to sound similarly, and therefore are more likely to be confused in spelling.

\subsection{Pseudo data for training}
Next we begin to make spelling perturbation to words.
Given a word $w$, we randomly select a source character $c$ from word $w$ and replace it with a target character $\tilde{c}$ drawn from the following the distribution:
\begin{equation}
    p_{substitute}(c \to \tilde{c}) = M(c, \tilde{c})
\end{equation}


When $\tilde{c}$ is a sequence of character that contains $c$, an insertion error is made, and vice versa for a deletion error. Hence, insertions, deletions are substitutions are all included in the process.

For a large number of correct documents, we randomly add the spelling errors to generate pseudo training data and train the nested RNN model by minimizing the cross-entropy loss with the perturbed sentences as inputs and the original sentences as outputs.

\section{Experiments}

\begin{table*}
\centering
\small
\begin{tabular}{l|ccc|ccc}
    \Xhline{1.0pt}
    \textbf{System} & \multicolumn{3}{c|}{\textbf{Development}} & \multicolumn{3}{c}{\textbf{Test}}  \\
                                         & P & R & $\textrm{F}_{0.5}$     & P & R & $\textrm{F}_{0.5}$  \\
    \hline
    enchant                              & 49.46 & 53.47 & 50.22                   & 53.40 & 58.33 & 54.32 \\
    scRNN         ~\cite{scRNN:2017:AAAI}& 62.64 & 50.46 & 59.76                   & 66.40 & 56.08 & 64.04 \\
    LSTM-Char-CNN ~\cite{charcnn}        & 66.95 & 55.32 & 64.25                   & 66.58 & 58.33 & 64.75 \\
    nested RNN                           & \textbf{71.10} & 56.94 & \textbf{67.73} & \textbf{71.77} & 61.26 & \textbf{69.39} \\
    \Xhline{1.0pt}
\end{tabular}
\caption{\label{results} Performance of spelling error correction on the JFLEG development and test set.}
\end{table*}

\subsection{Training Data}
We use words from a public dictionary and train the character-level translations model from words to phonetic symbols.
After the model converges, we obtain about 1000 character pairs with non-zero confusion coefficient by setting $\theta_{att}=0.2$.
The One Billion Word Benchmark corpus ~\cite{billion:2013} is used to generate pseudo data.
We make pseudo errors on words with a probability of $P_{err}=0.05$, which is selected according to performance on development set.

\subsection{Evaluation and baselines}
~\citet{Brill:2001} used a 10,000-word corpus of common English spelling errors in their paper, but the dataset is not available to us.
Other datasets in this field, such as those proposed by Roger Mitton \footnote{http://www.dcs.bbk.ac.uk/~ROGER/aspell.dat} and Peter Norvig\footnote{http://norvig.com/ngrams/spell-errors.txt}, only consider isolated errors and contain no context information.
Therefore, we build a new one based on JFLEG ~\cite{JFLEG}, a fluency-oriented corpus for developing and evaluating GEC.
One annotator corrects the spelling errors of the source sentences in JFLEG and leave other parts unchanged.
480 and 432 spelling errors are annotated for the develop set and test set, respectively.
Following ~\citet{conll:2014}, the annotations are made in $\textit{.m2}$ format and the $\textrm{F}_{0.5}$ score computed by \textit{MaxMatch} (M2) scorer ~\cite{m2:2011} is used as evaluation metric.

We evaluate our model in comparison to PyEnchant\footnote{https://github.com/rfk/pyenchant}, scRNN ~\cite{scRNN:2017:AAAI} and a modified LSTM-Char-CNN ~\cite{charcnn}.

The LSTM-Char-CNN was originally proposed for language modeling~\cite{charcnn} and later fused in a GEC model~\cite{schmaltz:2016:BEA11}.
We modify it for spelling error correction by letting it predict the current corrected word instead of the next word in language modeling task.
We reimplement LSTM-Char-CNN and scRNN and train them with the same pseudo data.

\subsection{Training details and results}

For the nested RNN, the character embedding size is set to 64. The hidden unit sizes of CharRNN and WordRNN are 256 and 512.
The CharCNN in LSTM-Char-CNN has 256 filters, 64 for each of the width [1,2,3,4].
The character vocabulary size and word vocabulary sizes are 84 and 30k for the 3 neural network models.

We train the models using Adam ~\cite{adam:2014} for 50k steps with $lr=0.001, \beta_1=0.9, \beta_2=0.999, \epsilon=10^{-8}$ and then switch to vanilla SGD with $lr=0.2$ and halve the learning rate every 50k steps.
We train a total of 200k steps with mini-batchsize of 96, which takes 20 hours on a single Tesla K40m GPU for the nested RNN.
We evaluate the models on development set every 10k steps and choose the best checkpoints for test.

Table~\ref{results} shows the performance of the models on the development and test set.
As seen, all the neural network methods outperform enchant by a large margin and the proposed nested RNN substantially improves upon other methods.

\section{Analysis}

\begin{table}
\centering
\small
\begin{tabular}{l|l}
    \Xhline{1.0pt}
    source           & I \textbf{though} that was right .  \\
    reference        & I \textbf{thought} that was right . \\
    \hline
    enchant          & I \textbf{though} that was right .  \\
    scRNN            & I \textbf{though} that was right .  \\
    LSTM-Char-CNN    & I \textbf{thought} that was right . \\
    nested RNN       & I \textbf{thought} that was right . \\
    \Xhline{1.0pt}
    source           & No matter they are \textbf{smell} or big . \\
    reference        & No matter they are \textbf{small} or big . \\
    \hline
    enchant          & No matter they are \textbf{smell} or big . \\
    scRNN            & No matter they are \textbf{small} or big . \\
    LSTM-Char-CNN    & No matter they are \textbf{smell} or big . \\
    nested RNN       & No matter they are \textbf{small} or big . \\
    \Xhline{1.0pt}
\end{tabular}
\caption{\label{example} Examples of real-word errors.}
\end{table}

Table~\ref{example} shows two examples of real-word errors where the nested RNN outperforms other models.
The enchant detects errors mostly based on dictionary looking up, and doesn't perform well in these cases.
The scRNN ignores the order of internal characters and relies heavily on the first and last characters to recognize a word, causing its failure to correct the word ``though'' to ``thought'' since the last character would change.

The last point we would emphasize is that the LSTM-Char-CNN resembles the nested RNN except that it uses a CharCNN to represent a word instead of a CharRNN.
This difference is crucial in that CNN is ``rigid'' in finding insertions and deletions errors.
Here we give a intuitive explanation.
For example, given a character sequence pattern $[abcd]$, the corresponding 4-gram kernel matching it is also $[abcd]$.
If we delete the character ``$b$'', the source pattern becomes $[acd]$, where the longest sequence that can be matched is $[cd]$, whose length is only $1/2$ of the original one.
In contrast, for a CharRNN, $3/4$ of the pattern remains unchanged in this condition since it reads characters sequentially.
Hence, the nest RNN is more suitable for correcting insertion errors and deletion errors than LSTM-Char-CNN.






\bibliography{acl2018}
\bibliographystyle{acl_natbib}

\end{document}